\documentclass{ecai}
\usepackage{graphicx}
\usepackage{latexsym}
\usepackage{inconsolata}
\usepackage{times}  
\usepackage{helvet}  
\usepackage{courier}  
\usepackage{url}  
\usepackage{makecell}
\usepackage{graphicx}  
\usepackage{amsmath}
\usepackage{amsfonts,amssymb}
\usepackage{enumerate}
\usepackage{caption}

\begin{document}

\begin{frontmatter}

\title{Tuning In to Neural Encoding: Linking Human Brain and Artificial Supervised Representations of Language}

\author[A]{Jingyuan Sun\thanks{Corresponding Author. Email: jingyuan.sun@kuleuven.be}}
\author[B]{Xiaohan Zhang}
\author[A]{Marie-Francine Moens}

\address[A]{Department of Computer Science, KU Leuven, Belgium}
\address[B]{Institute of Neuroscience, Key Laboratory of Primate Neurobiology, CAS Center for Excellence in Brain Science and Intelligence Technology, Chinese Academy of Sciences, China}

\begin{abstract}
To understand the algorithm that supports the human brain's language representation, previous research has attempted to predict neural responses to linguistic stimuli using embeddings generated by artificial neural networks (ANNs), a process known as neural encoding. However, most of these studies have focused on probing neural representations of Germanic languages, such as English, with unsupervised ANNs. In this paper, we propose to bridge the gap between human brain and supervised ANN representations of the Chinese language. Specifically, we investigate how task tuning influences a pretained Transformer for neural encoding and which tasks lead to the best encoding performances. We generate supervised representations on eight Natural Language Understanding (NLU) tasks using prompt-tuning, a technique that is seldom explored in neural encoding for language. We demonstrate that prompt-tuning yields representations that better predict neural responses to Chinese stimuli than traditional fine-tuning on four tasks. Furthermore, we discover that tasks that require a fine-grained processing of concepts and entities lead to representations that are most predictive of brain activation patterns. Additionally, we reveal that the proportion of tuned parameters highly influences the neural encoding performance of fine-tuned models. Overall, our experimental findings could help us better understand the relationship between supervised artificial and brain language representations.
\end{abstract}

\end{frontmatter}

\section{Introduction}

Neural encoding for language has long been a critical topic of interest in the intersection of natural language processing and cognitive neuroscience \cite{anderson2016predicting,holdgraf2017encoding}. The recent development of deep learning and pre-trained language representations \cite{sun-etal-2020-distill,sun-etal-2018-memory} has provided new opportunities  for enhancing the precision and efficiency of neural encoding \cite{sun2020neural,wang2022fmri}. Employing pre-trained language representations in neural encoding enables a more profound capture of the intricate relationships between linguistic stimuli and neural responses \cite{caucheteux2021model}. This, in turn, may foster the creation of superior language processing models \cite{schrimpf2021neural}. Additionally, probing the relationship between artificial and neural representations of language can yield valuable insights into the fundamental mechanisms underlying language processing \cite{rosenthal2018mapping}.

Despite the extensive research on unsupervised embeddings for English language in neural encoding \cite{ruan2016exploring, vodrahalli2017mapping}, there is a lack of studies exploring the use of supervised embeddings for neural encoding in other languages, such as Chinese \cite{wang2022synchronized}. Moreover, even the few studies that do adopt supervised embeddings for neural encoding in English still rely heavily on fine-tuning pre-trained models for task supervision \cite{oota-etal-2022-neura}. However, fine-tuning has been shown to distort the knowledge learned from pre-training \cite{kumar2022finetuning}, which is inconsistent with the human brain's mechanism that does not require a major reformation of the brain's language network to learn a single new task. To address these gaps, this paper proposes the use of both fine-tuned and prompt-tuned supervised sentence embeddings to fit a neural encoding model for Chinese. Prompt-tuning, which protects pre-trained knowledge by freezing weights and learning additional embeddings to fit a task \cite{li-liang-2021-prefi}, has yet to be widely explored for neural encoding, and this paper is to address this gap.

In pursuit of this goal, we employ both partial and full fine-tuning as well as prompt-tuning to adapt the pre-trained language model to 8 different NLU tasks, individually. The aim is to discern the influence of task tuning on a Transformer model for neural encoding and identify which tasks result in the best encoding performance. We find that: 
\begin{enumerate}[(i)]
\item  Prompt-tuning on 5 of the 8 tasks yields supervised representations that significantly exceed fully fine-tuned peers in predicting brain activities in the language network\footnote[1]{A brain functional network is a collection of brain regions that consistently show coordinated activity for certain cognitive functions or during behavioral tasks.}.  However, on none of the 8 tasks do fine-tuned embeddings significantly outperform the prompt-tuned ones in neural encoding
\item Tuning on tasks that require a compositional understanding of entities and concepts yields supervised representations that are better at neural encoding than tuning on other tasks. 
\item The proportion of tuned parameters highly influences the neural encoding performance of fine-tuned models. 

\end{enumerate}

In summary, this paper makes a triple-folded contribution. First, we propose a novel neural encoding framework with prompt-tuned supervised representations.
We prove it to be a viable alternative to fine-tuning-based methods. Second, we demonstrate how different tuning methods influence a pre-trained Transformer in neural encoding with comprehensive experiments. Third, our findings indicate that the balance between protecting pre-trained knowledge and learning task-related features is important for optimal neural encoding performance.  Overall, this work could help us better understand the relationship between task-tuned artificial and brain language representations.

\section{Related Work}

\subsection{Neural Encoding with ANN representations}

There has recently been a surge of discussions regarding the feasibility of ANNs as computational candidates for brain language representations due to their promising experimental results in NLU tasks \cite{caucheteux-etal-2021-model-base,schrimpf2021neural, sun2019towards}. 
In essence, previous studies have employed ANN representations to effectively predict brain responses to linguistic stimuli. Early research has proved their capability to predict brain activation patterns in response to simple verbal stimuli, such as individual words and phrases \cite{handjaras2016concepts,huth2016natural}. Subsequent studies have further displayed their ability to predict and decipher the representations of more complex stimuli \cite{anderson2016predicting,sun2020neural, sun2023contrast}.

Previous studies have primarily focused on the use of unsupervised ANN representations optimized for language modeling or similar context-predicting goals, where supervised models have been paid relatively limited attention \cite{sun2023ijcai}. A related study by Oota et al., \cite{oota-etal-2022-neura} utilized fine-tuning of a pre-trained BERT model on downstream tasks to compare neural encoding performance. However, full fine-tuning generally updates the entire set of pre-trained parameters and is distinct from the way the brain adapts to new tasks \cite{kumar2022finetuning}. It raises the question as to whether fine-tuning is the most appropriate framework for obtaining supervised representations for probing the human brain. In light of this, we undertake a comprehensive comparison between fine-tuned and prompt-tuned representations for neural encoding. 

Moreover, previous studies mainly targeted Germanic languages especially English, while Tibetan languages such as Chinese are under exploration. Partially limited by the scarcity of large-scale public brain imaging datasets for the Chinese language. 
Wang et al., (\cite{wang2022synchronized}) open-sourced a multi-modal neuroimaging dataset scanned from 12 subjects listening to 5 hours of naturalistic Chinese stories. We will base our experiments on this dataset. Wang et al., (\cite{wang2022synchronized}) also conducted basic neural encoding experiments mainly to evaluate the quality of the collected dataset but they did not employ any supervised representations.

\subsection{Tuning Pre-trained Models}

 \textbf{Fine-tuning.} \ The recent decade has witnessed a significant advancement in the development of pre-trained language models, yielding remarkable improvements in performance across a wide spectrum of NLP tasks \cite{qiu2020pre}. Fine-tuning is one of the most commonly employed techniques for task supervision which involves updating the pre-trained parameters to fit a specific task. Although fine-tuning typically results in favorable task performance, the computational time and space complexity is high 
 due to the need of calculating gradients and saving optimizer states for all parameters \cite{gu-etal-2022-pp}. Additionally, full fine-tuning has been proved to distort the general domain knowledge learned during pre-training, potentially undermining its cognitive plausibility \cite{Kumar2022FineTuningCD}.

\noindent\textbf{Discrete prompting.} \  Discrete prompting, as an alternative approach for fine-tuning, undergoes rapid development especially in the NLP field \cite{10.1145/3560815}. This technique leverages a pre-trained model's parameters to solve NLP tasks by querying the model with a discrete natural language prompt \cite{brown2020language}. For example, to conduct emotion categorization, a sample such as "I love this lovely kitten" could be prepended with a prompt like "Categorize the emotion". The pre-trained language model would be asked to predict the subsequent tokens. Despite its advantage in not requiring training and only storing a single set of model parameters,  discrete prompting's task performance may be highly impacted by the design of the prompt template and not be as optimal as fine-tuning in certain scenarios \cite{lester-etal-2021-powe}. Therefore we do not choose discrete prompting as a baseline approach in this study.

\noindent\textbf{Prompt-tuning.} \ Prompt-tuning builds upon discrete prompting. It fixes the pre-trained parameters of a  model and trains additional prompt embeddings to guide the model to learn a downstream task \cite{qin-eisner-2021-learnin}. 
Prompt-tuning has been demonstrated in previous research to exceed discrete prompting on various tasks (as reported in \cite{chen2022knowprompt}) and deliver task performance comparable to fine-tuning. Prefix-tuning \cite{li-liang-2021-prefi} is one typical implementation of prompt-tuning. It prepends trainable embeddings called "prefixes" to the original input word embeddings and optimizes them. 
In this study, we employ the P-tuning V2 method \cite{liu-etal-2022-pr}, an optimized version of prefix-tuning that compares or even surpasses fine-tuning on GLUE \cite{wang-etal-2018-glu}.

\section{Methods}

In the following subsections, we will first introduce how fine-tuning and prompt-tuning work. We then demonstrate how the tuned representations are used to build a neural encoder. 

\begin{figure}[ht]
\centering
\small
\includegraphics[width=3.4in]{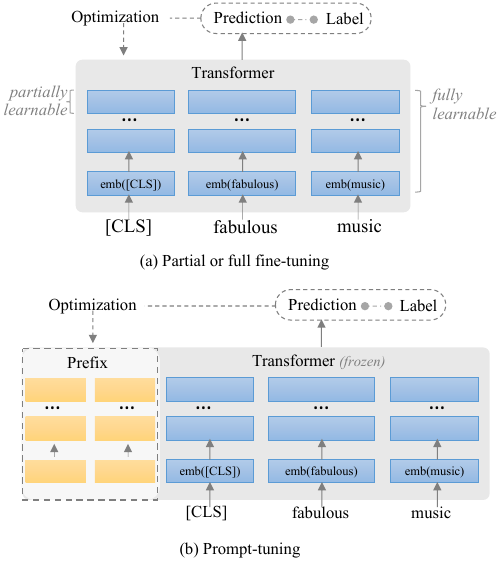}
\caption{Tuning a Transformer on a downstream task, taking sentiment classification as an example: [a] with partial or full fine-tuning; [b] with prompt-tuning.}
\label{fig_tune}
\end{figure}
\subsection{Tuning pre-trained Models}

This subsection introduces fine-tuning and prompt-tuning, using the example of tuning a Transformer-based language model for a conditional generation task. The language model, denoted by $LM = p_{\phi}(y|x)$, takes an input context $x$ and generates an output sequence of tokens $y$. The model is parameterized by $\phi$, and $z=[x;y]$ represents the concatenation of $x$ and $y$, with $id_{x}$ and $id_{y}$ denoting the corresponding sequences of indices. At each time step $i$, the activations of all hidden layers are represented as $h_i=[h_i^1, h_i^2, ... h_i^n]$, where $h_i^j$ denotes the activation of the $j$-th Transformer layer at time step $i$, and $h_i$ is the concatenation of all activation layers. The activations $h_i$ are computed as a function of $z_i$ and the past activations in their context, given by $h_i = \mathbf{LM}_{\phi}(z_i, h)$. 

To compute the distribution for the next token, the model uses the hidden states $h_i^{(n)}$ from the last layer and applies a softmax function to the pre-trained matrix $W_\phi$ to generate the logits for the vocabulary. Specifically, the distribution for the next token is computed as $p_\phi\left(z_{i+1} \mid h_{\leq i}\right)=\operatorname{softmax}\left(W_\phi h_i^{(n)}\right)$, where $h_i^{(n)}$ denotes the last layer hidden states of $h_i$. This process enables the model to generate coherent and contextually-relevant sequences of tokens.

\subsubsection{Fine-tuning}

As shown in Figure 1, fine-tuning is based on a pre-trained set of parameters $\phi$.
During fine-tuning, a language model distribution $p_\phi$ is trained by optimizing the following log-likelihood objective:
\begin{equation}
\max _\phi \log p_\phi(y \mid x)=\sum_{i \in id_{\mathrm{y}}} \log p_\phi\left(z_i \mid h_{<i}\right)
\end{equation}
 The fine-tuning methods can be further divided into partial and full fine-tuning.  Partial fine-tuning freezes part of pre-trained parameters $\phi$ and optimizes the remaining parameters such as these in 
 the layers near the output layer. Full fine-tuning optimizes the full set of pre-trained parameters.

\subsubsection{Prompt-tuning}

We build upon the P-tuning method \cite{liu-etal-2022-pr} to prompt-tune a pre-trained model on downstream tasks. In Figure \ref{fig_tune}, we illustrate the mechanism of P-tuning with a Transformer-based language model. P-tuning prepends a continuous trainable prefix to different layers of an ANN to guide its fitting on downstream tasks.

We now explain the mechanism of P-tuning following \cite{li-liang-2021-prefi}.
As shown in Figure 1, during P-tuning, a prefix, denoted as $Pr$, is prepended to the input sequence, resulting in the concatenation $z = [Pr; x; y]$. $Pr$ is a continuous trainable embedding, its corresponding sequence of indices is denoted as $Pr_{id_{x}}$ with length $|Pr_{id_{x}}|$. A trainable matrix $M_\theta^{Pr}$, with dimensions $|Pr_{id_{x}}| \times dim(h_i)$, is initialized to store the prefix parameters. And
\begin{equation}
h_i=\left\{\begin{array}{ll}
M_\theta^{Pr}[i,:], & \text { if } i \in {Pr}_{\mathrm{id_x}} \\
\mathbf{LM}_\phi\left(z_i, h_{<i}\right), & \text { otherwise. }
\end{array}\right.
\end{equation}
Training P-tuning is then to optimize the log-likelihood objective as in (1) with $\phi$ being fixed and $\theta$ being trainable.

\begin{figure}[ht]
\centering
\small
\includegraphics[width=3.3in]{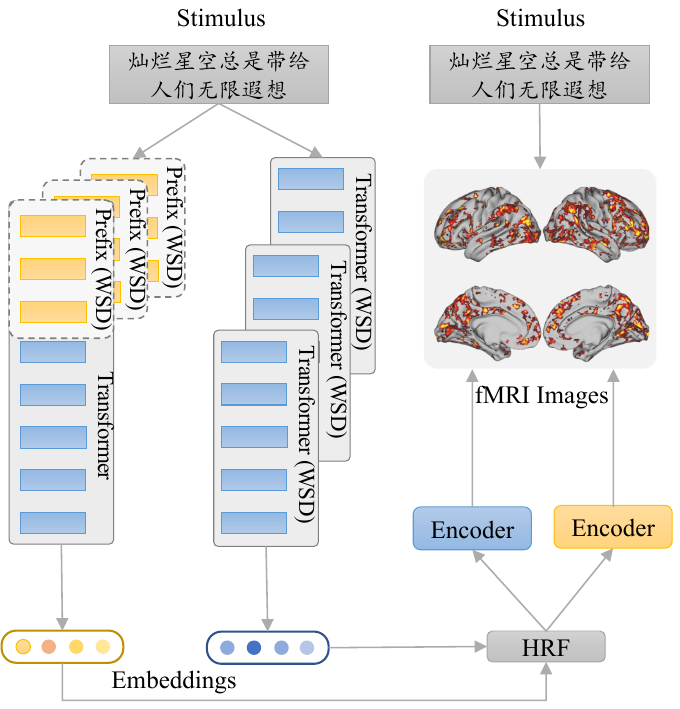}
\caption{Building neural encoders with fine-tuned and prompt-tuned supervised language representations. The example stimulus is taken from the fMRI dataset that we use in this work and its English translation is "The brilliant starry sky always gives people endless reverie".}
\label{fig_main}
\end{figure}

\begin{figure*}[h!]
\centering
\small
\includegraphics[width=5.5in]{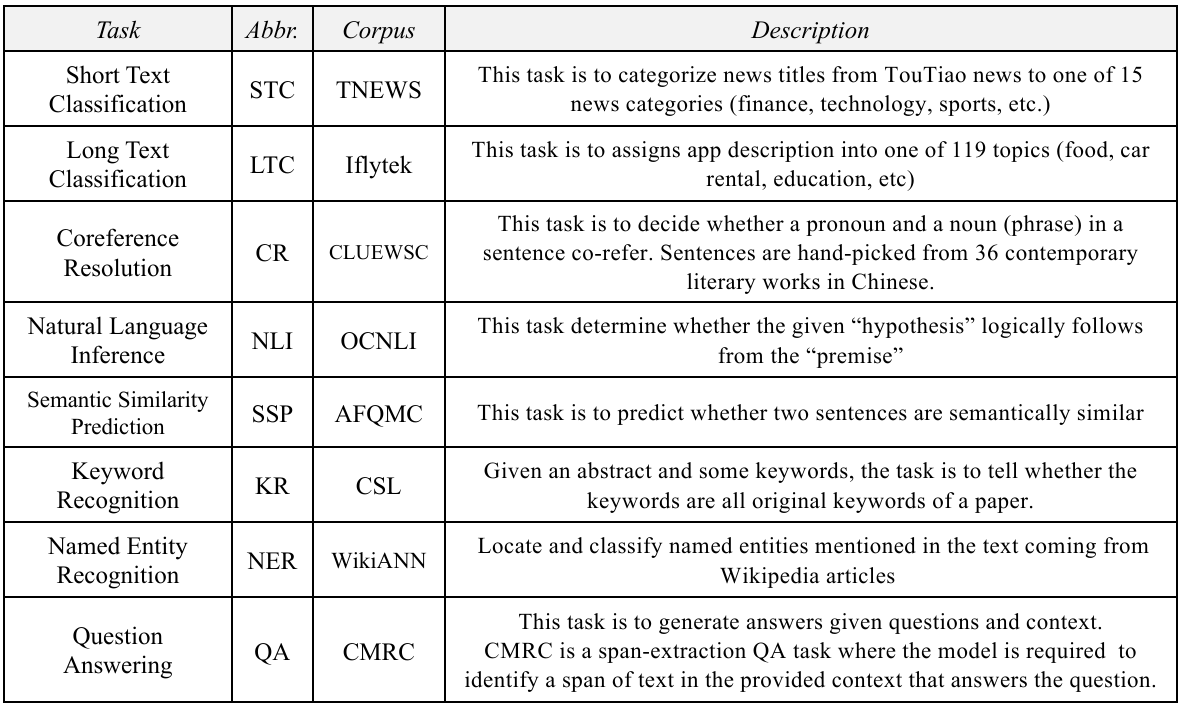}
\caption*{\textbf{Table 1}: 8 tasks for tuning and their abbreviations, corpus and descriptions. }
\label{tab2}
\end{figure*}
\subsection{Neural Encoder}

To learn the mapping between computational models and brain activation, we train voxel-wise neural encoding models to predict fMRI signals from sentence stimuli for each of the 12 participants.  The fMRI methodology measures the fluctuations in blood-oxygen-level-dependent (BOLD) signals, which undergo slow alterations post neuronal firing. To accommodate for the temporal delay between sentence stimuli and corresponding neural activity, the initial step involves the convolution of sentence embeddings with the canonical hemodynamic response function (HRF). The HRF delineates the transformation in the BOLD signal following neuronal firing.
Then, the convolved features were used to predict fMRI signals with Ridge regression.  

Specifically, within the training set, we have a voxel matrix $X_e\in \mathbb{R}^{N_E \times N_V}$ and a convolved sentence embedding matrix $Z_e\in \mathbb{R}^{N_E \times N_D}$. \(N_E\) signifies the number of examples, \(N_V\) represents the number of voxels, and \(N_D\) stands for the number of dimensions for sentence representation. The goal is to estimate the regression coefficients of the encoder $W_e$ by minimizing the following objective function for each column \(x_i\) in \(X_e\), which corresponds to the BOLD signal of each voxel:
\begin{align}
\begin{split}
|| W_eZ_e -x_i||_2^2+\lambda ||W_e||_1
\end{split}
\end{align}
Here, \(\lambda\) is the regularization parameter.

During the training process, the encoder is optimized using a 5-fold cross-validation procedure. This procedure is applied to data from each of all 12 subjects, with each kind of supervised sentence embedding as a separate input. To assess the performance of the encoder, we calculate the Pearson correlation between the predicted and measured fMRI signals.

\section{Experimental Setup}

In this section, we will first briefly introduce the brain imaging dataset we use. We then demonstrate the principle of task selection and some characteristics of the selected tasks. Code implementations are available online \footnote[2]{\url{https://github.com/soinx0629/sup_fmri_enc_chn/}} .

\subsection{Brain Imaging Data}

The Chinese fMRI dataset used in our experiment is from \cite{wang2022synchronized}.  It is one of largest existing fMRI datasets for naturalistic Chinese language narratives. In this study, 12 native Chinese speakers underwent scanning with a Siemens Prisma 3T scanner equipped with a 64-channel receiver coil while they were engrossed in listening to a collection of 60 Chinese stories. These stories covered a broad range of topics, with each story extending from 4 to 7 minutes, culminating in approximately 5 hours of audio content.   The text and audio of all stories were downloaded from Renmin Daily Review website\footnote[3]{\url{https://www.ximalaya.com/toutiao/30917322/}}, after which the text was manually checked to ensure its concordance with the audio content. There were 52,269 words in all stories, forming a vocabulary of 9,153 words. Subsequent to its collection, the fMRI data were preprocessed following the Human Connectome Project (HCP) pipeline \cite{glasser2013minimal}. The fMRI dataset is publicly available\footnote[4] {\url{https://openneuro.org/datasets/ds004078}}.

\begin{figure*}[ht]
\centering
\small
\includegraphics[width=6.75in]{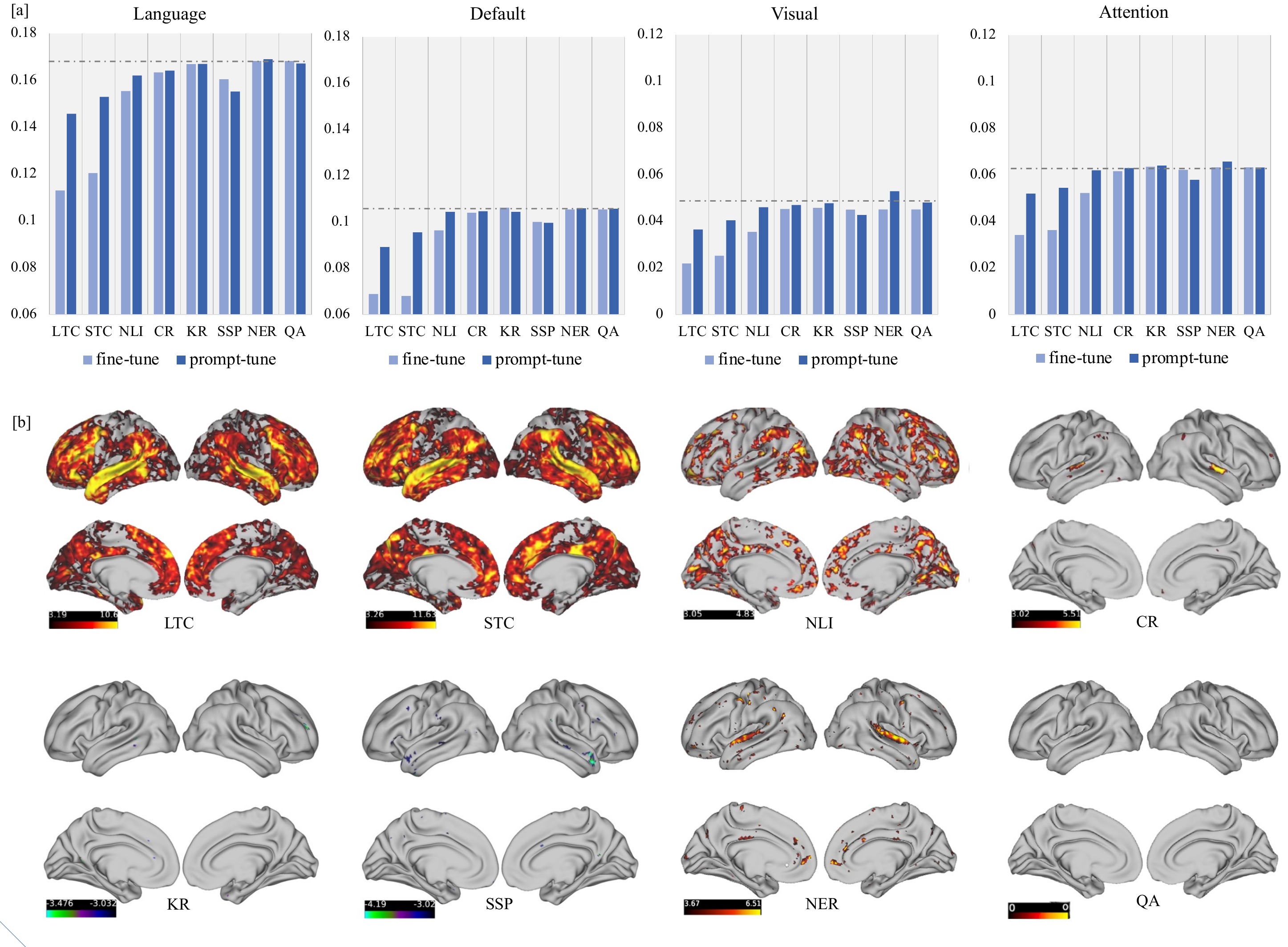}
\caption{[a] Neural encoding performance for the brain language network, DMN, visual network and the dorsal attention network. Dashed lines denotes the performance of the un-tuned RoBerTa model. The Y-axis denotes the performance in Pearson's correlation. [b] Significance results for comparing the encoding performance of prompt-tuned against fine-tuned representations. Pairwise t-test with FDR correction is conducted and  significance values are projected onto the inflated (top,
lateral and medial views) cortical surface.}
\label{fig_cor}
\end{figure*}

\subsection{Tasks for Tuning}

We select tasks for this work based on two major principles that have also been adopted by previous work \cite{oota-etal-2022-neura}. First, tasks that require a diverse set of cognitive-linguistic skills are chosen. Second, tasks from in widely used NLP benchmarks for example CLUE \cite{xu-etal-2020-clue} are desired.
The following 8 tasks are chosen for evaluation considering these principles, including
Long Text
Classification (\textit{LTC}), 
including Short Text
Classification  (\textit{STC}),  Natural Language Inferencing (\textit{NLI}), Co-reference Resolution (\textit{CR}), Keyword Recognition (\textit{KR}), Semantic Similarity Prediction (\textit{SSP}), Named Entity Recognition (\textit{NER}) and Question Answering (\textit{QA}). 
We present the details of these datasets shown in Table 1, including the task description and the corpus used.

\subsection{Sentence Representations}

In this study, we adopt the pre-trained RoBertTa-wwm-ext \cite{cui-etal-2020-revisiting} model\footnote[5]{https://huggingface.co/hfl/chinese-roberta-wwm-ext}. It is an improved version of naive RoBerTa model specially designed for Chinese. It achieved impressive performance on the CLUE benchmark as the base model. 
As shown in Figure 2, we adapt the RoBertTa with fine-tuning and prompt-tuning respectively on each of the 8 Chinese NLU tasks. 
Supervised representations for neural encoding are then generated by these tuned models.
Specifically, we feed each of sentence stimuli presented for scanning the fMRI images to the tuned models and average the hidden states of the last layer to represent the sentence. 
The generated sentence embeddings have a dimension of 768.
Both fine-tuned and prompt-tuned models are trained by ourselves on Nvidia 3080TI GPUs with 12GB VRAM. We conduct a hyper-parameter search to achieve optimal task performance.

\section{Experimental Results}

\subsection{Prompt-tuned against Fine-tuned Representations}

In this subsection, we compare the prediction performance with prompt-tuned and fully fine-tuned representations. 
We build neural encoders with task-tuned representations to predict brain activities within the language network, visual network, dorsal attention network, and default mode network. 
We conduct a group-level paired t-test with FDR correction to check the significance of comparison results. The results are shown in Figure \ref{fig_cor}.

As demonstrated in Figure \ref{fig_cor}a and \ref{fig_cor}b, prompt-tuning on 5 of the 8 tasks, namely LTC, STC, NLI, CR and NER, yields representations that exceed fine-tuned peers in predicting brain language network.
The most significant advantages of prompt-tuning representations over fine-tuned ones are achieved by tuning on the two classification tasks LTC and STC, where prompt-tuned embeddings significantly outperform in predicting the left superior temporal gyrus (LSTG), the dorsal area of left inferior frontal gyrus (LIFG) and middle frontal gyrus (LMFG).
LSTG, LMFG, and LIFG contain acknowledged important components of the brain language network \cite{fedorenko2011functional}. LMFG exhibits support for a dedicated "visual word form" system with orthographic specificity, which undergoes specialization to effectively represent orthographic knowledge \cite{hirshorn2016decoding}. LIFG is observed to facilitate verbal selection and possibly plays a domain-general role in the selection process for competing representations \cite{fedorenko2014reworking}.
Tuning on the NLI task also yields prompt-tuned representations that exceed fine-tuned peers in predicting the left and inferior parietal lobule, especially the rostrodorsal and caudal area.
Prompt-tuned CR representations surpass mainly in predicting the Superior Temporal Gyrus (STG). STG is an essential component of the brain's language network \cite{yi2019encoding}. It is particularly important for extracting meaningful linguistic features from speech input. 
Except for these 4 tasks above, prompt-tuning on the NER leads to higher predicting correlation on some parts of the language and auditory network. 
These comparison result stands up to the significance test as depicted in Figure  \ref{fig_cor}b.

Prompt-tuning on KR and QA tasks does not generate representations that significantly outperform fine-tuning.
Specifically, the neural encoding performance of prompt-tuned and fine-tuned task representations do not differ significantly in predicting more than 95\% of the cortical voxels.
Only tuning 1 of 8 tasks, the SSP task, produce fine-tuned representations outperforming prompt-tuned ones on predicting some micro potions of the brain language network, such as small subregions within the left and right STG.
The pattern of predicted performance is generally consistent across 4 functional networks. For example, not only within the language network but within the other 3 functional networks we also observe that prompt-tuned LTC, STC, and NLI representations significantly outperform fine-tuned ones in encoding correlation. Prompt-tuning on the other 4 tasks does not lead to significantly different performance from fine-tuning when predicting the major ROIs of the other 3 functional networks.

\begin{figure*}[ht]
\centering
\small
\includegraphics[width=6.85in]{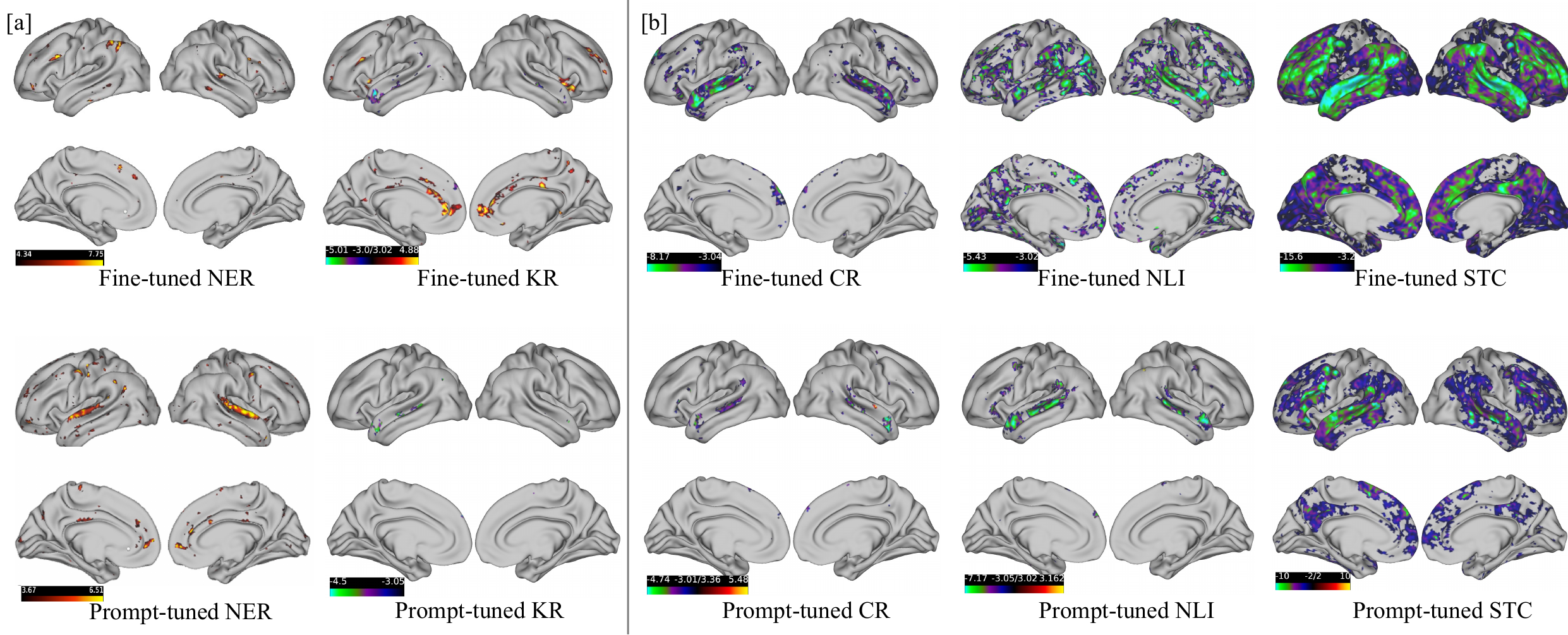}
\caption{Significance results for comparing the encoding performance of prompt-tuned and fine-tuned representations against the untuned RoBerTa model. A pairwise t-test with FDR correction is conducted and significance values are projected onto the inflated cortical surface. Prompt-tuning on the two tasks in [a] significantly improves over the untuned model in some ROIs. Tuning on the three tasks in [b] underperforms the untuned model in encoding the major ROIs in brain language networks.}
\label{fig4}
\end{figure*}
\begin{figure*}[ht]
\centering
\small
\includegraphics[width=6.81in]{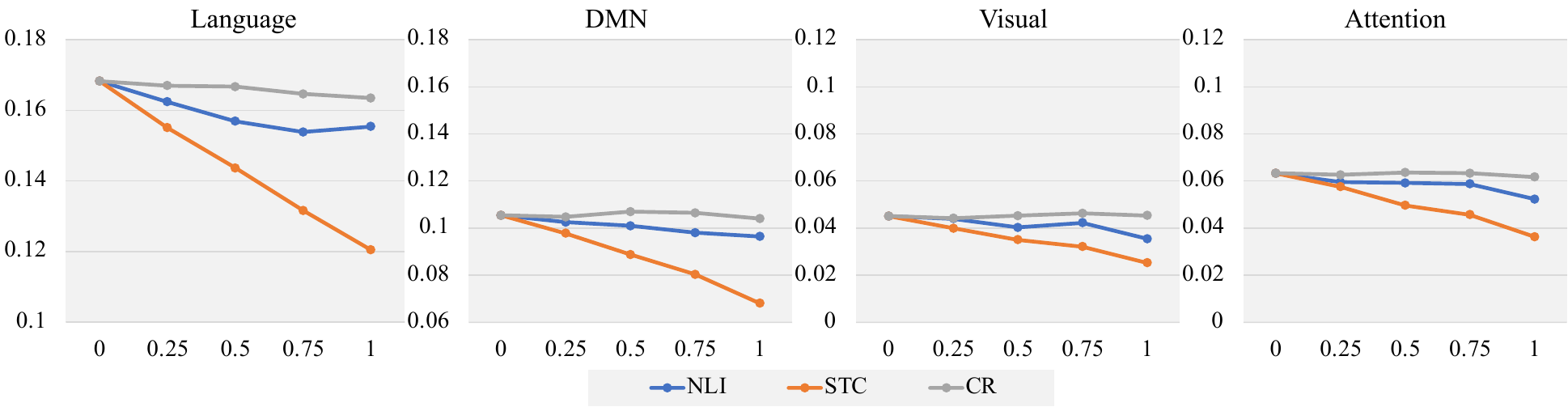}
\caption{Influences of tuned parameter proportion on a fine-tuned model's neural encoding performance. The X-axis denotes the tuned proportion and the Y-axis denotes the encoding performance in Pearson's correlation.}
\label{fig5}
\end{figure*}

\subsection{Tuned against Untuned Representations}

In the previous subsection, we compare how prompt-tuned and fine-tuned representations are in neural decoding. It is a logical next step to ask how the tuned and untuned naive representations differ in predicting neural responses. We have depicted the neural encoding performance of the un-tuned model in 4 functional networks in Figure 3[a].
In this subsection, we go deeper to compare the prompt-tuned and fine-tuned representations respectively with the untuned model in predicting all cortical voxels. The significance of comparison results is checked by group t-test with FDR correction and depicted in Figure 4.

We find that only tuning on 2 of the 8 tasks significantly improves over the untuned model. As shown in Figure 4[a], tuning on NER and KR yields neural encoding performance that significantly exceeds untuned models on predicting ROIs of the language and auditory network. 
Prompt-tuning on the NER task surpasses the untuned model in encoding the left and right superior temporal gyrus (STG) especially the A4 and A5 area. It also surpasses in area 55b which is a part of dorsal auditory stream.
Fine-tuning on the KR task improves over the untuned model in predicting the subgenual area of the cingulate gyrus (CG) and medial area of the right superior frontal gyrus (RSFG). 
Both KR and NER tasks require fine-grained compositional processing of entity and concept.
So these empirical results might implicate that fine-grained representation of entities and concepts support the formation of brain language representation of sentences.

Fine-tuning and prompt-tuning on the NLI, STC, CR, SSP, and LTC task yield significantly worse neural encoding performance. Prompt-tuning exerts less influence on encoding performance than fine-tuning.
We select 3 of the 5 tasks to display in Figure 4[b]. 
We find that tuning on these tasks decreases the neural encoding performance in major ROIs of the brain language network. 
Both fine-tuning and prompt-tuning on CR, NLI, and STC tasks significantly decrease neural encoding performances on STG.
Tuning on STC further impairs the encoding performance on large potions of the middle frontal gyrus and Inferior parietal lobule, especially the rostrodorsal areas. 
 Not only in the language network, NLI, and STC task tuning also significantly impairs encoding correlation in DMN and visual networks.  The findings indicate that pre-trained knowledge distorted by fine-tuning contributes more to neural encoding than the task-related features learned for these 5 tasks.

\subsection{Influences of Tuned Parameter Proportion}

In previous subsections, we find that prompt-tuned representations of 5 tasks significantly outperform fine-tuned ones in encoding the brain language networks. We also demonstrate that even in the tasks where both fine-tuning and prompt-tuning decrease the encoding correlation compared to the untuned model, the influence of prompt-tuning is significantly lower than full fine-tuning. We analyze that one major difference between prompt-tuning and full fine-tuning is the proportion of tuned parameters. Prompt-tuning freezes all the pre-trained weights and learns additional embeddings to fit on a task, while full fine-tuning re-optimizes all the pre-trained weights. 

In this subsection,  we investigate whether the proportion of tuned parameters influences the encoding performance of fine-tuned representations. We partially fine-tune the pre-trained RoBerTa on the  tasks found in the last subsection that lead to significantly worse encoding performance than the un-tuned model. We set the tuned parameter proportion to be 0.25, 0.5, and 0.75 respectively, and train neural encoders with these partially fine-tuned task representations. Altogether with the un-tuned and fully fine-tuned model, we depict the neural encoding performance in Figure 5.

As demonstrated in Figure 5, the tuned parameter proportion highly influences the encoding performances of fine-tuned representations, especially in the brain language network. Taking the classification task STC as example, we find a significantly positive correlation $(p<0.01)$ between the encoding performance with the tuned parameter proportion. With the increase of the tuned proportion, the encoding correlations of STC rapidly drop. A tuned parameter proportion of 1, which is fully fine-tuning, leads to the worst neural encoding performance.
Not only in the language network but in the DMN and attention network, we also observe significant positive correlations between the tuned parameter proportion and encoding performance (all significant results have $(p<0.01)$).

In summary, the findings of this subsection inform that the proportion of tuned parameters highly influences the encoding performance of fine-tuned representations. Especially for tasks on which tuning decreased the neural encoding performances, the more parameters tuned, the worse neural encoding performances go.

 \section{Discussion}

 In the previous section, we compare the encoding performance of prompt-tuned against fine-tuned representations, as well as the encoding performance of tuned representations against the untuned naive model. 
 We also demonstrate the most predictive task representation on each cortical voxel. 
 In this section, we provide further analysis and possible explanations for the experimental results.

 First, we find that tuning on 5 of the 8 tasks yields prompt-tuned representations that exceed fine-tuned peers in predicting voxels within the brain language network. 
 One major difference between fine-tuning and prompt-tuning is their treatment of the pre-trained weights. 
 To learn a new task, fine-tuning generally updates the entire parametric space of a pre-trained model while prompt-tuning freezes the pre-trained weights and optimizes additional task-specific embeddings. 
 Moreover, fine-tuning has been proven to distort the pre-trained features. 
So intuitively prompt-tuning can better protect the general domain knowledge acquired from pre-training than fine-tuning methods. 
Then the different encoding performances of fine-tuning and prompt-tuning can be at least partially explained by the importance of general domain knowledge in predicting the brain activations. If the general domain knowledge is more important than the task-related features for encoding the brain activities, then we can expect tuning methods that better protect pre-trained knowledge yields higher encoding correlation.

Second, we find that KR and NER representations more accurately predict brain activations in most cortical regions than other task-supervised representations, particularly in the regions of interest (ROIs) that compose the brain's language network.
The KR task is to gauge a model's ability to determine whether the abstract can be summarized by the keywords.  The NER task tests a model's ability to identify and classify named entities in text into predefined categories.
The NLI and CR task representations rank second and third in prediction correlation. 
All these tasks require a compositional understanding of entities, concepts, and their semantic relations.
This implies the possibility that capturing fine-grained concept and entity features supports the human brain in building representations for linguistic units.

Third, we investigate the impact of the proportion of tuned parameters on the encoding performance of fine-tuned representations. We suggest that the proportion of tuned parameters significantly affects the encoding performance of fine-tuned representations, particularly for tasks where tuning decreases the neural encoding performance. The more parameters tuned, the worse the neural encoding performances becomes. 
Notably, fully fine-tuning the model leads to the worst neural encoding performance. A larger tuned proportion means more pre-trained knowledge might be distorted by fine-tuning. So these findings indicate the importance of knowledge learned in Transformer pretraining for neural encoding. It is necessary for the supervised models to strike a balance between the protection of pre-trained knowledge and learning task-related features to achieve optimal neural encoding performances.

\section{Conclusion}

In this paper, we aim to link the representation of the Chinese language in supervised artificial neural networks (ANNs) and the human brain. Our investigation focuses on task tuning and its influence on the neural encoding of a pre-trained Transformer. To achieve this, we utilize prompt-tuning, a technique that is rarely explored in neural encoding for language, to generate supervised representations on 8 Natural Language Understanding (NLU) tasks. Our findings reveal that prompt-tuning leads to better representations for predicting neural responses to Chinese stimuli compared to traditional fine-tuning on 5 tasks. Interestingly, we observe that representations of tasks that requires fine-grained processing of concepts and entities are most predictive of brain activation patterns. We also discover that the proportion of tuned parameters significantly influences the neural encoding performance of fine-tuned models. In summary, our experimental findings provide insights into the relationship between supervised ANNs and brain language representations and could pave the way for future investigations into language representations in the brain.

\section*{Acknowledgements}
This work is funded by the CALCULUS project (European Research Council Advanced Grant H2020-ERC-2017-ADG 788506).

\bibliography{ecai}
\end{document}